\documentclass[11pt]{article}

\usepackage[final]{acl}

\usepackage{times}
\usepackage{latexsym}

\usepackage[T1]{fontenc}

\usepackage[utf8]{inputenc}

\usepackage{microtype}

\usepackage{inconsolata}

\usepackage{graphicx}

\usepackage[colorinlistoftodos]{todonotes}

\newcommand{\cready}[1]{} 

\usepackage{enumitem}
  \setlist{itemsep=0.5pt, topsep=0pt,leftmargin=1.4em,labelsep=0.5em}

\usepackage{hyperref} 
\usepackage{xcolor}
\hypersetup{
    colorlinks,
    linkcolor={red!50!black},
    citecolor={blue!50!black},
    urlcolor={blue!80!black}
}

\usepackage{amsmath}
\usepackage[capitalize,nameinlink]{cleveref}
\Crefname{equation}{Eq.}{Eqs.}
\Crefname{figure}{Fig.}{Figs.}
\Crefname{tabular}{Tab.}{Tabs.}

\usepackage{etoolbox} 
\makeatletter
\pretocmd{\appendix}{%
  \@addtoreset{figure}{section}%
  \@addtoreset{table}{section}%
}{}{}
\makeatother

\usepackage{bbm} 

\usepackage{amsmath}
\usepackage{amsmath, amssymb, mathtools}
\usepackage{booktabs}
\usepackage{multirow}
\usepackage{tabularx}
\usepackage{booktabs}   
\usepackage{float}      
\usepackage{dblfloatfix} 

\title{
Robustness as an Emergent Property of Task Performance
}

\author{Shir Ashury-Tahan$^{\spadesuit}$, Ariel Gera$^{\spadesuit}$, Elron Bandel$^{\spadesuit}$, \\
\textbf{Michal Shmueli-Scheuer$^{\spadesuit}$ and Leshem Choshen$^{\spadesuit\heartsuit\clubsuit}$}\\
$^{\spadesuit}$IBM Research $^{\heartsuit}$MIT $^{\clubsuit}$Weizmann Institute of Science\\}

\begin{document}
\maketitle
\begin{abstract}

Robustness is widely viewed as a key challenge for real‑world applications. However, because current research focuses only on difficult tasks, it partially captures real-world readiness.
In this paper, we argue and verify that robustness -- consistency across semantically equivalent inputs -- closely follows task difficulty: once models master a task, robustness emerges naturally.
Through an empirical analysis of multiple models across diverse datasets and configurations (e.g., paraphrases, temperature changes), we observe a strong positive correlation between task performance and robustness.
Furthermore, our findings indicate that robustness is driven primarily by task-specific competence rather than inherent model attributes, challenging the common view of robustness as an independent capability.
This perspective implies that as tasks mature and model performance saturates, robustness on those tasks will similarly emerge.
For researchers, this suggests that explicit efforts to measure robustness may deserve reduced emphasis, as robustness is likely to improve alongside performance. 
For practitioners, it signals that while many existing benchmarks are still unstable, models are already reliable on earlier tasks and suitable for deployment.


\end{abstract}

\begin{figure}
    \vspace{-0.6cm}

    \centering
    \includegraphics[width=0.88\linewidth]{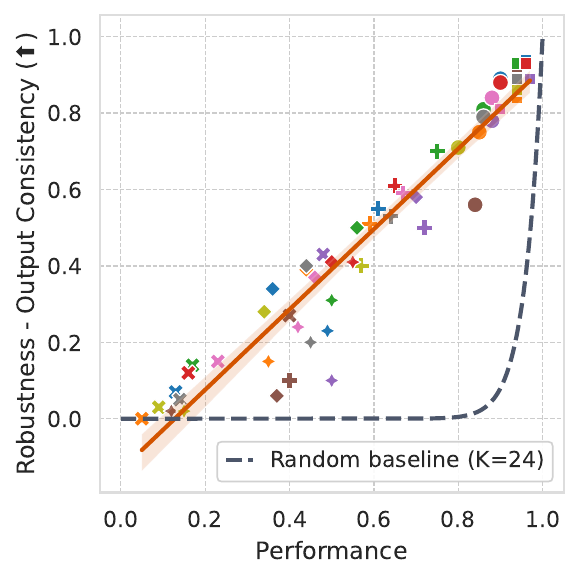}
    \caption{Linear regression of robustness on performance.
    Robustness increases slightly faster than performance, with a slope of $1.05$. Performance explains $92.4\%$ of robustness variance, indicating a strong predictive relationship.
    Colors denote models, and shapes represent datasets. Dashed gray line: random baseline, i.e., the probability of answering consistently across all example configurations, assuming per-configuration success probability equals the model’s performance.}
    \label{fig:robust_vs_saturation_lin_reg}
    \vspace{-0.6cm}
\end{figure}

\section{Introduction}

\begin{quote}
\footnotesize
    \emph{One man's trash is another man's treasure}
\end{quote}


Robustness -- the ability of models to produce consistent outputs across semantically equivalent inputs -- becomes increasingly critical as AI scales into high-stakes applications \citep{kumar2025robustnesslargelanguagemodels, guo2023evaluatinglargelanguagemodels}. Even state-of-the-art models often exhibit unpredictable behavior under minor input changes, undermining confidence in their reliability and hence practicality, especially in scenarios where stability is critical \citep{Yang2024AssessingAR, Wang2024RUPBenchBR, ashurytahan2025mightytorrbenchmarktable}.

Most robustness works focus on challenging tasks and not on saturated tasks \citep{voronov-etal-2024-mind, pezeshkpour2023largelanguagemodelssensitivity,sclar2024quantifyinglanguagemodelssensitivity,alzahrani2024benchmarkstargetsrevealingsensitivity, ashurytahan2025mightytorrbenchmarktable, mizrahi2024stateartmultipromptllm, kumar2025robustnesslargelanguagemodels}. Thus, they are not able to capture a full picture of real‑world readiness, where model behavior on saturated tasks is relevant for deployment decisions.
Moreover, these works evaluate robustness as a distinct signal, separate from performance, assuming it to be an inherent property of the model.
We challenge this view by proposing that robustness depends on task difficulty, rather than model skill alone; a lack of robustness in one setting does not indicate global brittleness. Missing evaluations on saturated tasks obscure this link and leave a key aspect of robustness hidden.

In this work, we examine robustness across different levels of task difficulty, considering it not as a standalone property but in relation to model performance. We focus on a key dimension of real‑world robustness: stability under \textit{meaning‑preserving input variations}.
We argue and verify that \textbf{performance serves as a meaningful signal of model robustness}. When models demonstrate high success and questions are unquestionably easy for them, they are easy regardless of how they are presented.
To explore this hypothesis, we analyze multiple models across diverse datasets and configurations, spanning a range of difficulties. 
We conduct dedicated experiments as well as analyze results from prior benchmarks.

Our findings reveal a strong positive correlation between benchmark performance and model robustness, demonstrating that as model performance approaches the upper limits of a task, so does its resilience to inference variations. As evident from the random‑baseline robustness in Figure~\ref{fig:robust_vs_saturation_lin_reg}, this phenomenon goes well beyond the ``trivial robustness'' expected from high success rates, and remains consistent across diverse model architectures. 

Importantly, our results reveal that robustness is not predominantly an intrinsic property of the model, as often assumed in prior work. Instead, we find that \textit{task‑specific competence is the main driver of robust behavior}: models exhibit high robustness primarily on tasks they have mastered, regardless of architecture or scale.




Our findings have implications for both practitioners and researches.
For real‑world deployment, the results suggest that extremely strong task performance is a reliable indicator of a model’s consistency on that task, supporting its readiness for safe use.
For researchers, the results indicate that robustness may warrant less focus, as robustness appears to be a concomitant effect that increases as a benchmark approaches saturation. Thus, as model saturation extends to new tasks over time, robustness on those tasks is likely to arise naturally.

\begin{figure*}

    \centering
    \includegraphics[width=0.95\linewidth]{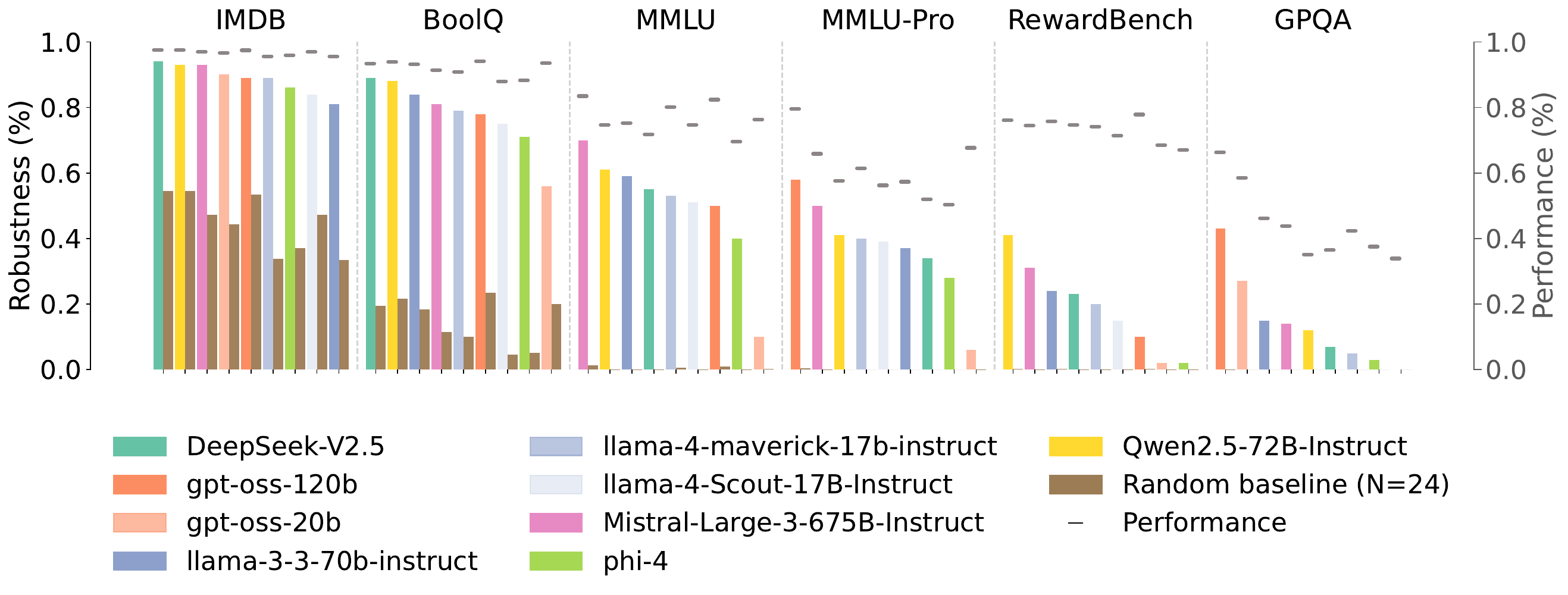}
    \caption{
    Robustness rate per dataset, computed using the output consistency metric, in relation to overall performance (represented by dashed lines). Robustness rises as benchmarks saturate. Camel bars represent the random baseline consistency probability with similar success rates, which is approximately $0$ when performance is below \textasciitilde$80\%$. Additional robustness metrics analysis can be found in App.~\S\ref{sec:append_results}. 
    }
    \label{fig:std_all_models}
    \vspace{-0.5cm}
\end{figure*}

\section{Preliminaries}\label{sec:preliminaries}

Let $D$ denote a dataset, with examples $\{x_i \mid i\in \{1, \dots, |D|\}\}$.
Each example can be inferred using one of the configurations $v \in V$, denoted as $x_i^v$. 


Let $m(x_i^v)$ represent the model prediction for $x_i^v$, and let $\mathrm{score}(m(x_i^v))$ denote the benchmark evaluation score assigned to the prediction.

In what follows, we define the key terms used throughout the experiments we conducted.\footnote{These definitions are closely related to those used in the ToRR benchmark, which is analyzed in Section~\ref{sec:results}. However, they differ in certain technical details, including the set of configurations considered and the procedure used to aggregate robustness.}
    
    \paragraph{Inference Configuration}
    Configurations are chosen to reflect  naive user inputs in real‑world settings of diverse types, for which the model is expected to produce identical outputs. 
    This includes surface form variations (paraphrasing), in-context modifications (modifying demonstrations and their quantity), generation parameter changes (varying temperature), and prompt noise (such as random string addition). 
    Formally, given an example $x_i$, we define a set of configurations $\{x_i^v \mid v \in V\}$, which share the same semantic meaning but differ in how they are presented to the model. The detailed configuration can be found in App.~\S\ref{app:exp_setup}.

In our main results, all configurations are equally valid references and considered original.
Otherwise, we denote $x_i^o$ the original reference configuration. We calculate metric scores against each original reference and then average the results.
    
\paragraph{Model Capability}
The performance score (e.g., accuracy, F1) obtained by a model on the original version.
We define the overall capability of a model $m$ on dataset $D$ as the average score across all examples:
\[
\text{Capability}(m, D) = \frac{1}{|D|} \sum_{i=1}^{|D|} \mathrm{score}(m(x_i^o))
\]
\paragraph{Model Robustness (Output Consistency)}
Following previous work, we define robustness at the example level \citep{Nalbandyan2025SCORESC, ackerman-etal-2024-novel, zhu2024promptrobustevaluatingrobustnesslarge, habba2025dovelargescalemultidimensionalpredictions, ashurytahan2025mightytorrbenchmarktable} as strict agreement of model predictions across configurations.

Given the set of configurations $V$, a model $m$ is considered \emph{robust} on example $x_i$
if all configuration outputs are equivalent:
\[
c_i = \mathbf{1}\Big[ m(x_i^{v_1}) \equiv m(x_i^{v_2}) \;\; \forall v_1, v_2 \in V \Big]
\]
Dataset-level robustness is the fraction of examples that are robust:
\[
\text{OutputConsistency}(m, D)
\;=\;
\frac{1}{|D|} \sum_{i=1}^{|D|} c_i
\]

\section{Experimental Setup}

Our analysis combines dedicated experiments as well as results from existing benchmarks.

Our experiments span 9 models and 6 diverse datasets, covering difficulty levels from sentiment analysis (IMDB) and reading comprehension (BoolQ) to advanced reasoning (MMLU, MMLU‑Pro), reward‑model evaluation (RewardBench), and graduate‑level problem solving (GPQA). From each dataset, we sample $100$ examples and generate predictions under $24$ different configurations, resulting in roughly $130k$ total inferences. We then evaluate, for each model–dataset pair, performance\footnote{We validate against existing benchmarks (see App~\S\ref{app:perform_verify}).} and robustness across these configurations. More details are provided in App.~\S\ref{app:exp_setup}.

To complement these experiments, we analyze results from the ToRR benchmark \citep{ashurytahan2025mightytorrbenchmarktable}, an independent collection of $10$ diverse table reasoning datasets spanning multiple domains, tasks, and evaluation metrics. We selected this benchmark because it primarily consists of generative tasks, providing a complementary evaluation setting to our classification-focused experiments. Moreover, its robustness metric is defined at the example level, enabling consistency with our robustness analysis. We do not conduct any additional experiments on ToRR; instead, we leverage the benchmark's existing results to provide further evidence for the claims presented in our results section (\S\ref{sec:results}). Further details about ToRR are provided in App.~\S\ref{app:torr_analysis}.

\paragraph{Robustness Primary Metric}
To keep robustness and performance measurements as independent as possible, we use \textit{OutputConsistency} as our main robustness metric. 
Unlike performance, robustness is computed solely from model outputs and aggregated across example configurations.

\paragraph{Additional Robustness Metrics}
To further support our findings, we evaluate two additional score-based metrics: (i) the standard deviation of scores across example configurations, and (ii) the performance drop rate across configurations. Formal definitions are provided in App.~\S\ref{app:robust_metrics}.

\paragraph{Random Baseline}
We also compare our results to a random baseline, computed as the probability of consistent answers across all configurations, assuming that the success probability of each configuration equals the model’s overall performance.

\paragraph{Contamination}
While contamination is a potential concern in evaluation studies, we mitigate it by using \textit{diverse} datasets and configurations, reducing the likelihood of exact overlap with pretraining corpora. 
Moreover, our analysis emphasizes consistency in all $24$ varied configurations, making it unlikely that contamination alone explains the observed robustness patterns. 
As shown in App.~\S\ref{app:model_full_std_dists}, models do not favor any specific configuration; instead, consistency values follow a long‑tail distribution across datasets and models.


\section{Results}\label{sec:results}

This section presents our main finding: performance is strongly correlated with robustness. For the experiments we run, robustness is measured using \textit{OutputConsistency} (\S\ref{sec:preliminaries}), while for the ToRR benchmark we use the benchmark's leaderboard, which is based on its unique robustness metric (\S\ref{app:torr_analysis}). Additional results and statistical analyses are provided in App.~\ref{sec:append_results}.







\subsection{Higher Performance is Associated with a Greater Proportion of Consistent Answers}

Our analysis suggests that the ability to solve a dataset is also associated with the ability to generalize across configurations. For example, all models on IMDB achieve performance between $95\%\text{ and }97\%$, and their robustness is comparatively high, ranging from $81\%\text{ to }94\%$.
This is not trivial, as illustrated in Figure~\ref{fig:std_all_models}, and becomes evident when comparing consistency to a random baseline with a similar overall success rate, which in the case of IMDB case achieves only between $33\%\text{ and }54\%$. 

A similar pattern emerges in the ToRR benchmark. For example, WikiTQ is among the easiest datasets, with most models achieving scores in the $80\%$--$90\%$ range, and correspondingly high robustness scores of roughly $75\%$ (see App. Table~\ref{tab:wikitq_perf_robust_sorted}). In contrast, on FinQA, where model performance is substantially lower, typically around $40\%$, robustness is also reduced, with values around $50\%$ (see App. Table~\ref{tab:finqa_perf_robust_sorted}).

More broadly, these results indicate a strong positive relationship between task performance and robustness.
Figure~\ref{fig:robust_vs_saturation_lin_reg} shows a clear correlation trend, with performance explaining $92.4\%$ of the variance in robustness. Consistent with this observation, when analyzing results at the dataset level, we found a Pearson correlation of $0.91$ between robustness and aggregated performance scores on the ToRR benchmark.

\subsection{Model-Specific Factors are Comparatively Weaker}
Our analysis shows that although architecture and design matter, their effect on consistency is modest relative to the strong performance–robustness trend in Figure~\ref{fig:robust_vs_saturation_lin_reg}. Nevertheless, models can differ in inherent robustness; for example, \texttt{gpt-oss-120b} is far more robust than \texttt{gpt-oss-20b} on datasets where their performance is similar (e.g., on BoolQ, a 1-point performance difference results in 120B achieving $78\%$ robustness compared to $56\%$ for 20B; similarly, on MMLU, a 6-point performance gap corresponds to a $40\%$ robustness difference).

The ToRR benchmark provides additional evidence for this observation. Models with comparable table reasoning performance usually exhibit highly similar robustness scores, as can be seen in Appendix Table~\ref{tab:finqa_perf_robust_sorted}. Notably, this pattern holds both within model families and across different families. For example, \texttt{Claude-3.5-Sonnet} and \texttt{Gemini-1.5-Flash} achieve comparable performance scores ($43\%$ and $43\%$) and exhibit similar robustness ($56\%$ and $57\%$), despite belonging to different model families. Likewise, within the Llama family, \texttt{Llama-3.1-405B-Instruct} and \texttt{Llama-3.1-70B-Instruct} attain similar performance ($36\%$ and $37\%$) and robustness ($49\%$ for both) on the FinQA dataset.

\subsection{Models Outperform Random Baselines Across Benchmark Regimes}
Models substantially exceed the random baseline on all benchmarks\footnote{This analysis was conducted only for the experiments we ran ourselves. For ToRR, each dataset requires a dataset-specific random baseline function, which is not provided by the benchmark.}, underscoring robustness that is not explained by chance. 
For the four datasets on the right of Figure~\ref{fig:std_all_models}, the random baseline achieves zero robustness (as performance is lower than 80\%). Even in cases of very high success probability (IMDB, BoolQ), the gap remains significant, $43.4\%$ and $62.9\%$ on average, respectively. 
An extreme case is \texttt{llama-4-Scout-17B-Instruct} on BoolQ, which outperforms the random baseline by $70.4\%$.

\subsection{Perturbations Act as Arbitrary Noise}
Significance testing using ANOVA shows that our findings are not driven by particular experimental choices, with configuration parameters (e.g., replacing prompt space chars with TAB chars) exerting only a minor effect on performance (see Appendix~\S\ref{app_sec:anova}). 
This suggests that perturbations of different types are largely equivalent, and may reflect different ways of injecting noise into the generation task. 

Similarly, the ToRR benchmark reports no consistently dominant perturbation or table serialization strategy. Instead, performance differences are distributed across perturbation types, with no single perturbation emerging as systematically more challenging than the others. This suggests that robustness scores are not driven by sensitivity to any particular perturbation. Rather, the perturbations primarily act as different sources of noise that preserve the underlying task semantics.

Finally, our results show few outliers, indicating little contamination, otherwise, the original would often be high when its paraphrases are not.

\section{Related Work}

\paragraph{Robustness}
LLM robustness has been widely studied, with work highlighting brittleness across tasks
\citep{alzahrani2024benchmarkstargetsrevealingsensitivity}, domains \citep{ashurytahan2025mightytorrbenchmarktable}, input perturbation types \citep{mizrahi2024stateartmultipromptllm}, or robustness as a phenomenon and how to address it \citep{kumar2025robustnesslargelanguagemodels}. However, these works have treated robustness as an isolated phenomenon.
\citeauthor{lunardi2025robustnessreliabilitybenchmarkbasedevaluation} 
show that while model rankings remain stable under paraphrasing, absolute effectiveness often drops.
\citet{Ding2018OnTS} 
further link adversarial robustness to input distribution.
\paragraph{Saturation Progress}
Evaluation benchmarks for LLMs has become increasingly saturated \cite{bengio2025internationalaisafetyreport, Reuel2024BetterBenchAA, ai-benchmarks-hit-saturation, Ott2022MappingGD, bengio2025internationalaisafetyreport}. Studies show rapid early gains followed by plateaus as models near perfect scores, often accelerated by data overlap \citep{Sainz2023NLPEI}, underscoring the pace and capabilities of current systems.
Recent work addresses saturation by introducing harder benchmarks, weighted metrics, or progressively challenging evaluations \citep{ivanov2025resurrectingsaturatedllmbenchmarks, Mirzadeh2024GSMSymbolicUT, etzine-etal-2025-revitalizing, Bradley2024EnhancingLE}. 
While these works highlight challenges such as evaluation limits or propose solutions, other perspectives on what saturation implies for models remain underexplored.

\paragraph{Generalization and Learning Order}
Prior work in both NLP and vision shows that LLMs tend to acquire and generalize tasks in a consistent order across architectures and training regimes; that is, some tasks are systematically easier to learn than others
\citep{hacohen2020letsagreeagreeneural, pliushch2022deepclassifiersagreeanalyzing, baldock2021deeplearninglensexample, choshen2022grammarlearningtrajectoriesneurallanguage,edamadaka2025universallyconvergingrepresentationsmatter}.
Recent studies also reveal striking similarities in learned parameters \citep{kaushik2025universalweightsubspacehypothesis}.
Whereas previous studies primarily examine generalization to unseen examples after the task has been learned, we demonstrate that the same structured behavior applies to meaning‑preserving variations of those inputs, capturing an additional dimension of task competence.

\section{Discussion}

While model robustness is often studied in isolation, here we take a broader perspective. We find that robustness to semantically equivalent inputs is mainly tied to overall task performance, where high performance corresponds to robust and consistent model behavior. Interestingly, intrinsic signals from the model itself are relatively weak.

At a higher level, our work relates to the dynamics of saturation, where tasks become progressively easier for models over time. Together with the observed link between performance and robustness, it suggests that, without explicitly addressing it, \textit{as models advance, robustness may increasingly cease to be a primary bottleneck}. 

These patterns matter for applications where robustness is as critical as accuracy (e.g., healthcare, safety).
They suggest that high model performance may function as an empirical indicator of the model's readiness for reliable deployment. While this means that top current benchmarks likely do not measure tasks that can be used in sensitive domains, it also means that older tasks that were abandoned by the research community and never shown to be robust, are likely a possibility.

Our findings indicate that perceived model robustness often reflects task-specific competence rather than inherent model properties, calling for a reduced focus on measuring and improving robustness in isolation.

Given the strong signals observed, a natural future direction is to explore other robustness types (e.g., adversarial robustness) to see whether similar patterns arise.

\section*{Limitations}

\paragraph{Scope of Tasks} Our experiments focus on classification and generative tasks. There may exist other settings in which the observed relationship between performance and robustness is weaker or does not hold.

\paragraph{Model Behavior} 
Our analysis and conclusions are based on a diverse set of models. While the observed behaviors are consistent across this set, they may not generalize under substantial shifts in model architectures or training paradigms.

\paragraph{Evaluated Models}
Due to cost constraints, we did not include closed-source models in our evaluation. Consequently, caution should be exercised when generalizing these results to all model types.

\paragraph{Robustness Types}
Our evaluation focuses only on lightweight, semantic‑preserving prompt perturbations. Thus, our conclusions may not extend to other robustness settings, such as strong adversarial attacks, non–semantic-preserving shifts, or tool‑use scenarios.





\section*{Acknowledgments}
We thank Eliya Habba for valuable discussions that contributed to the development of the core idea and insights presented in this paper.

\bibliography{custom}

\null
\newpage
\appendix

\section{Experimental Setup} \label{app:exp_setup}

\subsection{Experimental Design}

\begin{table*}[t] 
    \centering
    \small 
    \setlength{\tabcolsep}{6pt} 

    \begin{tabularx}{\textwidth}{l c X}
        \toprule
        \textbf{Configuration Parameter} & \textbf{Number of variations} & \textbf{Values} \\
        \midrule
        Paraphrases             & 2 & (a paraphrase created with an LLM for each dataset) \\
        Number of demonstrations & 2 & 
        2, 4
        \\
        Random noise            & 3 & no noise; replace prompt spaces with another character (e.g., TAB); add a random string at the beginning and end (70 chars) \\
        Model temperature       & 2 & 
        0.2, 0.6
        \\
        \bottomrule
    \end{tabularx}

    \caption{Configuration parameters used in the experimental setup. We ran all experiments using the 24 unique configurations listed in the table. 
    }

    \label{tab:config_params}
\end{table*}

Building on the definitions above, we conduct experiments on both saturated and less saturated benchmarks, incorporating different configurations. 

\paragraph{The Data}
We conducted our experiments using the following benchmarks: IMDB \citep{maas-EtAl:2011:ACL-HLT2011}, BoolQ \citep{clark2019boolqexploringsurprisingdifficulty}, MMLU~\citep{hendrycks2021measuringmassivemultitasklanguage}, MMLU-Pro~\citep{wang2024mmluprorobustchallengingmultitask}, GPQA~\citep{rein2023gpqagraduatelevelgoogleproofqa}, and RewardBench~\citep{lambert2024rewardbenchevaluatingrewardmodels}.
The datasets were chosen to share a similar classification task and use the same exact-match accuracy metric, ensuring comparability. 

\paragraph{The Models}
For each dataset, we evaluated the capability and robustness of $9$ open-weight models and $6$ model families, all listed in Table~\ref{tab:benchmark_scores}. These models were selected as open-weight representatives from different model families, allowing us to analyze how performance on a benchmark relates to robustness behavior. Each model was evaluated using all perturbation types described below.

\paragraph{Configurations}
Following previous work \citep{habba2025dovelargescalemultidimensionalpredictions, mizrahi2024stateartmultipromptllm, alzahrani2024benchmarkstargetsrevealingsensitivity}, we implement our experiments using the following variations with exact parameter values provided in Table~\ref{tab:config_params}:
\begin{enumerate}
\item \textbf{Paraphrases}: An LLM judge paraphrased the original prompt, and the resulting text was used as an alternative template.
\item \textbf{Number of Demonstrations}: We varied the number of in-context examples provided.
\item \textbf{Random Noise}: Random patterns were added as prefixes, suffixes, or space replacements of varying lengths to introduce noise.
\item \textbf{Model Temperature}: Inference was performed under different temperature settings.
\end{enumerate}

In total, we apply $24$ configurations, each representing a unique combination of these variations to assess the robustness of model performance. The use of multiple configurations helps mitigate contamination concerns that could affect our results.

\paragraph{Evaluation}
Since our experiments focused on classification tasks, we evaluated results using exact match between the gold answer and the model output. 
We instructed the model to output the final answer only. Prior to output comparison, both strings were normalized (i.e., lowercasing, and stripping whitespace).

\paragraph{Required Computation}
We sampled \(100\) examples from each dataset and evaluated each model on all samples across \(24\) configurations. This yields \(14{,}400\) inferences per model and \(129{,}600\) total inferences across nine models.



\begin{table*}
\small
\centering 
\begin{tabular}{lcccccc}
\toprule
\textbf{Model 
} & \textbf{IMDB} & \textbf{BoolQ} & \textbf{MMLU} & \textbf{RewardBench} & \textbf{MMLU-Pro} & \textbf{GPQA}\\
\midrule
gpt-oss-120b & .97 & .94 & .82 & .78 & .80 & .66 \\
gpt-oss-20b & .97 & .94 & .76 & .69 & .68 & .68 \\
\addlinespace

Mistral-Large-3-675B-Instruct & .97 & .91 & .84 & .75 & .66 & .44 \\
\addlinespace

llama-4-maverick-17b-instruct & .96 & .91 & .80 & .74 & .62 & .42 \\
llama-3-3-70b-instruct & .95 & .93 & .75 & .76 & .57 & .46 \\
Llama-4-Scout-17B-Instruct & .97 & .88 & .75 & .71 & .56 & .34 \\
\addlinespace

Qwen2.5-72B-Instruct & .97 & .94 & .74 & .76 & .57 & .35 \\
\addlinespace

DeepSeek-V2.5 & .97 & .93 & .72 & .75 & .52 & .36 \\
\addlinespace

phi-4 & .96 & .88 & .70 & .67 & .50 & .38 \\
\midrule
\textbf{Average} & .97 & .92 & .76 & .73 & .61 & .45 \\

\bottomrule
\end{tabular}
\caption{Benchmark performance summary.}
\label{tab:benchmark_scores}

\end{table*}


\subsection{Performance Verification with External Benchmarks} \label{app:perform_verify}

To sanity-check our results, we compared the performance scores we have got and presented in Table~\ref{tab:benchmark_scores}
against published scores from external sources. A practical challenge is coverage:
our model list includes several newer models, whereas some benchmarks
(e.g., IMDB, BoolQ) are \emph{older} and are not consistently reported for recent models.
Accordingly, we focus our cross-check on widely reported tasks such as MMLU, MMLU-Pro and GPQA.
The results presented below indicate a positive correlation between the scores and only minor differences, despite variations in evaluation runs between our setup and theirs.

\paragraph{Llama models.}
We compared our measurements with the scores reported on the official model cards and community
evaluations (e.g., the Llama 4 Maverick model card on Hugging Face%
\footnote{\url{https://huggingface.co/meta-llama/Llama-4-Maverick-17B-128E-Instruct}}).
Overall, the numbers are closely aligned (Table~\ref{tab:llama_verify}).

\begin{table}[th]
    \small
    \centering
    \begin{tabular}{lcc}
        \hline
        \textbf{Model} & \textbf{MMLU} & \textbf{MMLU-Pro} \\
        \hline
        Llama~4 Maverick~17B & 80 \,(85) & 62 \,(62) \\
        Llama~4 Scout        & 75 \,(79) & 56 \,(58) \\
        Llama~3.1 70B        & 75 \,(79) & 57 \,(53) \\
        \hline
    \end{tabular}
    \caption{Our measured scores (outside parentheses) versus published scores (in parentheses).
    Minor differences are expected due to evaluation details (e.g., prompt templates, decoding settings).}
    \label{tab:llama_verify}
\end{table}

\paragraph{HELM Capabilities}
We used the HELM \citep{liang2023holisticevaluationlanguagemodels} capabilities leaderboard, which reports results for MMLU-Pro and GPQA on some of our reported models. The results are similar to ours (see Table~\ref{tab:helm_verify}). One difference that may explain their slightly better scores is that they ran the evaluation with chain-of-thought (CoT) prompting, whereas we requested a final answer only.

\begin{table}[th]
    \small
    \centering
    \begin{tabular}{lcc}
        \hline
        \textbf{Model} & \textbf{MMLU-Pro} & \textbf{GPQA} \\
        \hline
        Llama~4 Scout        & 75 \,(74) & 56 \,(50) \\
        Llama~4 Maverick~17B & 80 \,(81) & 62 \,(65) \\
        \hline
        Qwen2.5-72B-Instruct & 57 \,(63) & 35 \,(42) \\
        \hline
        gpt-oss-120b         & 80 \,(79) & 66 \,(68) \\
        gpt-oss-20b          & 68 \,(74) & 68 \,(59) \\
        \hline
    \end{tabular}
    \caption{Our measured scores (outside parentheses) versus published scores (in parentheses).
    Minor differences are expected due to evaluation details (e.g., running with CoT).}
    \label{tab:helm_verify}
\end{table}

\null
\newpage
\null
\newpage
\null
\newpage

\section{ToRR Benchmark Analysis} \label{app:torr_analysis} To further validate our findings, we analyzed the ToRR benchmark \citep{ashurytahan2025mightytorrbenchmarktable}. ToRR provides an independent evaluation setting, as it differs substantially from our benchmark in both data characteristics and task design. It comprises $10$ datasets spanning diverse domains and table reasoning capabilities, most of which involve generative tasks and employ different evaluation metrics. 

As in many robustness studies, the ToRR benchmark introduces its own robustness measure tailored to its setting. Notably, this metric is different from all robustness measures used in our experiments, being defined as the complement of the average score range per example. We view this distinction as beneficial, as it allows us to examine whether our findings hold across different robustness definitions rather than being tied to a specific metric. 


The fact that this relationship holds despite substantial differences in datasets, tasks, evaluation protocols, and robustness metrics provides additional evidence that our findings generalize beyond the specific setting considered in this work.

While this study focuses on robustness at a fine-grained level, supporting evidence for our claims can also be found at the aggregate level. In particular, one may consider the ToRR leaderboard's mean performance score and mean accuracy score, which reflect overall model capability across tabular tasks. The trends observed in these aggregate metrics are consistent with our findings, providing \textit{additional evidence} that the claims presented in this paper remain valid beyond the low-level analysis.

\begin{table}[t] \centering \small \begin{tabular}{lcc} \toprule Model & Performance & Robustness \\ \midrule Claude 3.5 Sonnet (20241022) & .92 & .80 \\ DeepSeek v3 & .91 & .78 \\ GPT-4o (2024-11-20) & .91 & .75 \\ Llama 3.1 Instruct (70B) & .91 & .76 \\ Llama 3.1 Instruct (405B) & .90 & .76 \\ Gemini 1.5 Pro (002) & .89 & .69 \\ Gemini 1.5 Flash (002) & .88 & .72 \\ GPT-4o mini (2024-07-18) & .88 & .73 \\ Qwen2 Instruct (72B) & .86 & .75 \\ Claude 3.5 Haiku (20241022) & .85 & .61 \\ Llama 3.1 Instruct (8B) & .80 & .67 \\ Mistral Instruct v0.3 (7B) & .68 & .40 \\ Mixtral Instruct (8x22B) & .67 & .39 \\ Mixtral Instruct (8x7B) & .62 & .29 \\ \bottomrule \end{tabular} \caption{WikiTableQuestions performance (F1) and robustness scores from the ToRR leaderboard} \label{tab:wikitq_perf_robust_sorted} \end{table}

\begin{table}[t] 
\centering 
\small 
\begin{tabular}{lcc} \toprule Model & Performance & Robustness \\ \midrule Gemini 1.5 Pro (002) & .47 & .59 \\ DeepSeek v3 & .46 & .51 \\ Claude 3.5 Sonnet (20241022) & .43 & .56 \\ Gemini 1.5 Flash (002) & .43 & .57 \\ GPT-4o (2024-11-20) & .40 & .55 \\ Qwen2 Instruct (72B) & .38 & .51 \\ Llama 3.1 Instruct (70B) & .37 & .49 \\ Llama 3.1 Instruct (405B) & .36 & .49 \\ Claude 3.5 Haiku (20241022) & .35 & .50 \\ GPT-4o mini (2024-07-18) & .34 & .47 \\ Mixtral Instruct (8x22B) & .28 & .47 \\ Mixtral Instruct (8x7B) & .20 & .49 \\ Mistral Instruct v0.3 (7B) & .19 & .51 \\ Llama 3.1 Instruct (8B) & .04 & .63 \\ \bottomrule 
\end{tabular} \caption{FinQA program accuracy and robustness scores from the ToRR leaderboard.} \label{tab:finqa_perf_robust_sorted} 
\end{table}

\null
\newpage

\section{Robustness Metrics}\label{app:robust_metrics}

\subsection{Robustness Main Metric}

While it is common practice to measure robustness using scores, we find it somewhat less ideal to reflect both performance and robustness using the exact same numbers.
Instead, we rely primarily on the model's outputs in the paper, which offer a more direct reflection of its ability to maintain consistent predictions under meaning-preserving variations. 

It is important to note that this choice \textit{does not affect the validity} of our results: all robustness metrics we examined yielded similar trends. 

\subsection{Other Robustness Metrics}

We used the following robustness metrics to complement our main output-based measure and provide a broader perspective on model behavior under different configurations:

\paragraph{Standard Deviation (STD)}
We compute the standard deviation of the model's scores across the original and perturbed versions of each input. Formally, for a given input $x_i$ and its variants $\{x_i^v \mid v \in V\}$, we calculate:
$$\sigma_{x_i} = \text{STD}\Bigg(
\left\{ \mathrm{score}(m(x_i^v)) \mid v \in V \right\}
\Bigg)$$

A value of $0$ indicates perfectly consistent behavior across perturbations, while higher values reflect increased variability in the model's responses.



\paragraph{Performance Drop Rate (PDR)}
We follow \cite{zhu2024promptrobustevaluatingrobustnesslarge}
and calculate the relative drop in performance when the model is evaluated on perturbed inputs compared to the original test set. We define:

$$\rho_{x_i}~(\%) = \frac{1}{|V|}\sum_{v \in V} [
    1 - 
    \frac{\mathrm{score}(m(x_i^o))}{\mathrm{score}(m(x_i^v))}
     ]$$
$$PDR~(m, D) = \frac{1}{|D|}\sum_{x_i \in D}
    \rho_{x_i}$$

Here, performance is measured using the primary evaluation metric. A higher percentage indicates a greater sensitivity to perturbations.


\null
\newpage

\begin{figure*}[b]
    \centering
    \includegraphics[width=\textwidth]{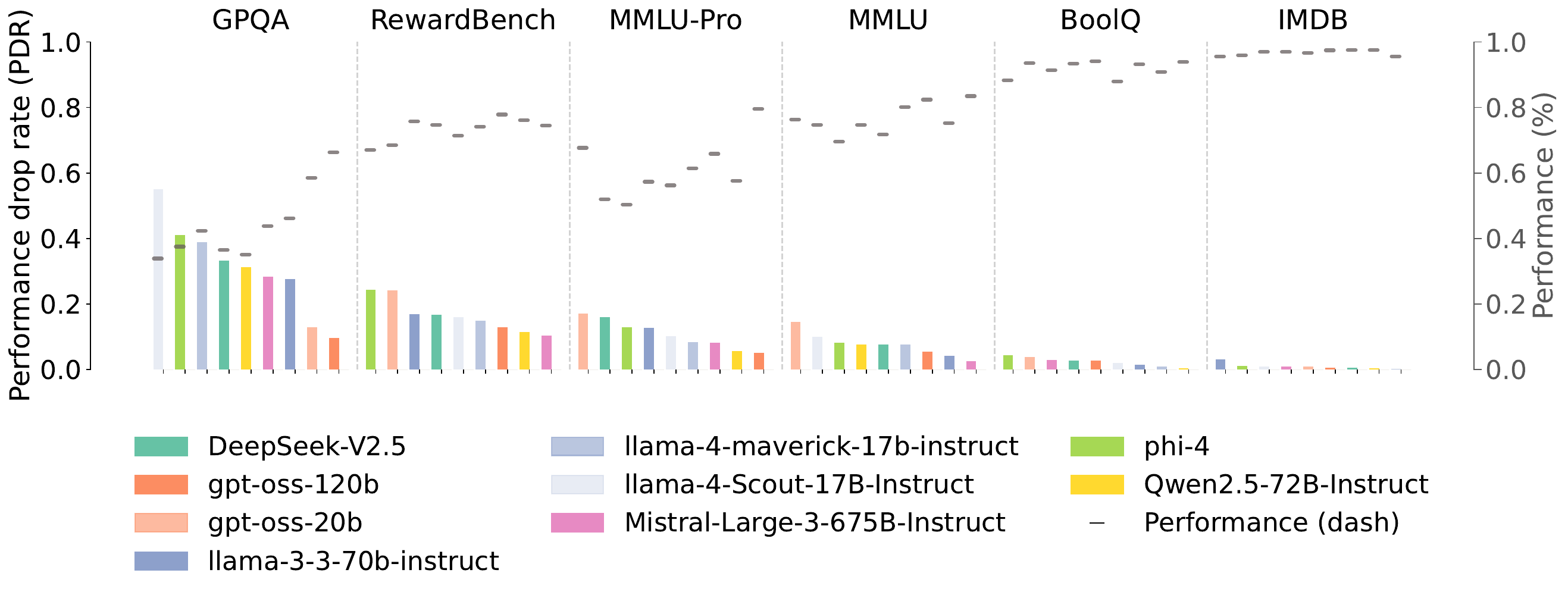}
    \caption{Average PDR reveals a similar trend: lower values indicate smaller performance drops across perturbations, signaling higher robustness and more consistent model behavior. The dashed lines indicates the performance.}
    \label{fig:pdr_all_models}
\end{figure*}

\null
\newpage

\section{Results}\label{sec:append_results}

We provide the results of  $2$ additional robustness metrics described in App.~\S\ref{app:robust_metrics}, and show they correlate well with our main results in the paper. While the primary metric used in the paper in output consistency which is string-based, these metrics provide another aspects that are score based.






\subsection{PDR Trends}

Calculating the PDR at the example level and averaging across models reveals a pattern similar to our main result as can be seen in Figure~\ref{fig:pdr_all_models}. Here, the bars illustrate how far performance can vary per instance, essentially indicating where score consistency is lacking, which complements what we examined in Figure~\ref{fig:std_all_models}, where the bars reflected output consistency. In this figure, they represent performance variability rather than stability. We observe the same dataset ordering as before, but reversed from left to right: GPQA remains the least robust and IMDB the most robust. The relationship with overall performance is also evident when comparing the bar trend to the dashed trend.




\begin{figure*}[b]
    \centering
    \includegraphics[width=\textwidth]{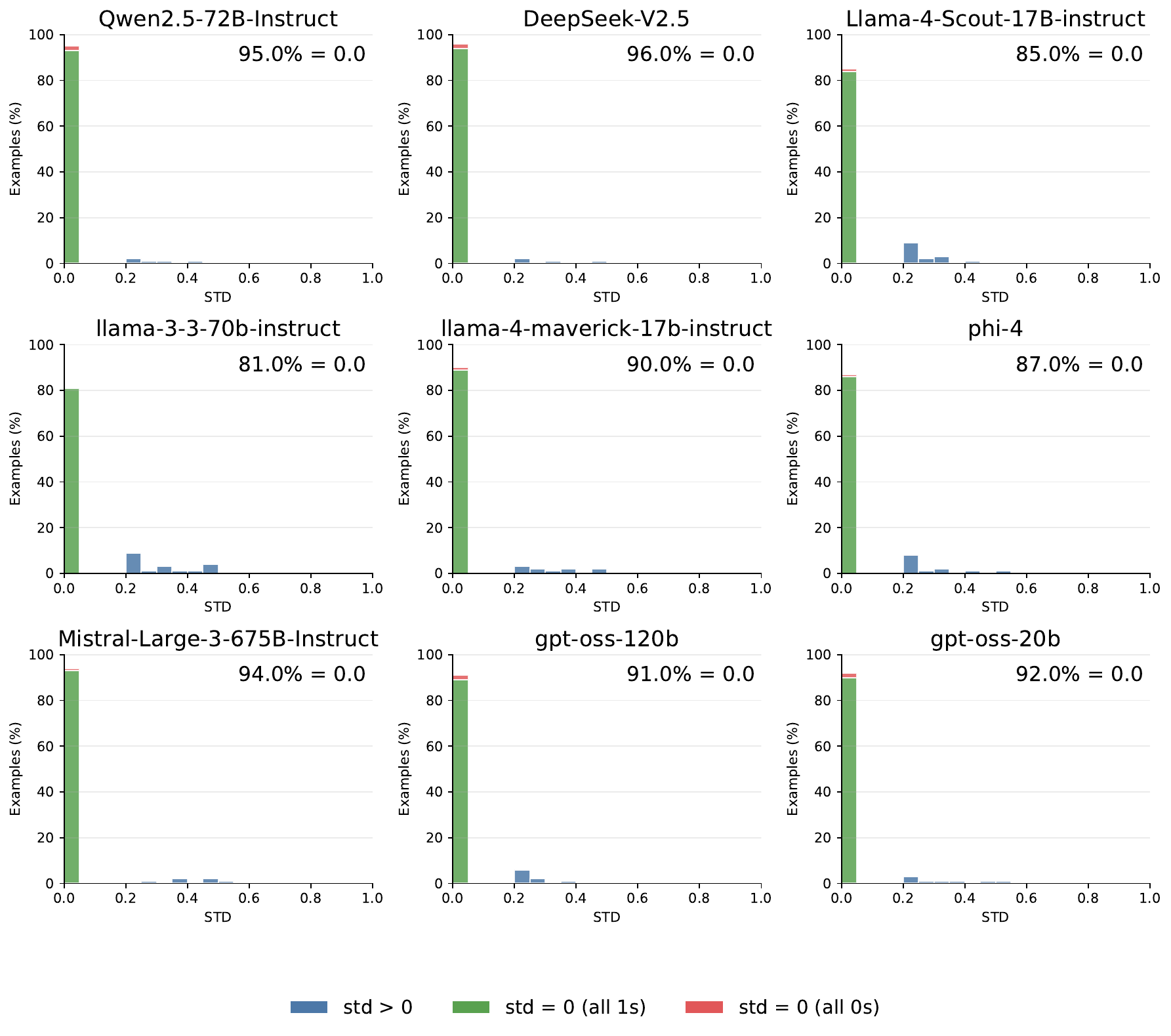}
    \caption{Per-example STD distribution for each model on the IMDb dataset.}
\end{figure*}


\subsection{Model-Dataset Full STD Distributions} \label{app:model_full_std_dists}

To provide a more granular perspective on our results, we extended the STD metric by calculating the full distribution of per-example STD across models. These distributions reveal detailed patterns of model consistency.

While an STD of $0$ indicates a consistent score, it is important to distinguish between success consistency and failure consistency. In our setup, success consistency means producing the same correct answer across all configurations, whereas consistent failure can occur due to different incorrect answers, which does not necessarily indicate robust behavior. Therefore, in the figures, we not only reflect the STD score but also differentiate whether an STD of $0$ corresponds to all failures or all successes.

Considering only successful cases, the results closely mirror what we presented in Figure~\ref{fig:std_all_models}. While that figure reflected output consistency, here we focus on score consistency. It is also evident that as robustness decreases, the STD distribution becomes flatter, whereas higher performance leads to an 
long-tail distribution.

It offers additional evidence that inherent model robustness is limited, as the STD distribution is similar across models within the same benchmark, i.e., their graph trends follow a comparable pattern.

\begin{figure*}
    \centering
    \includegraphics[width=\textwidth]{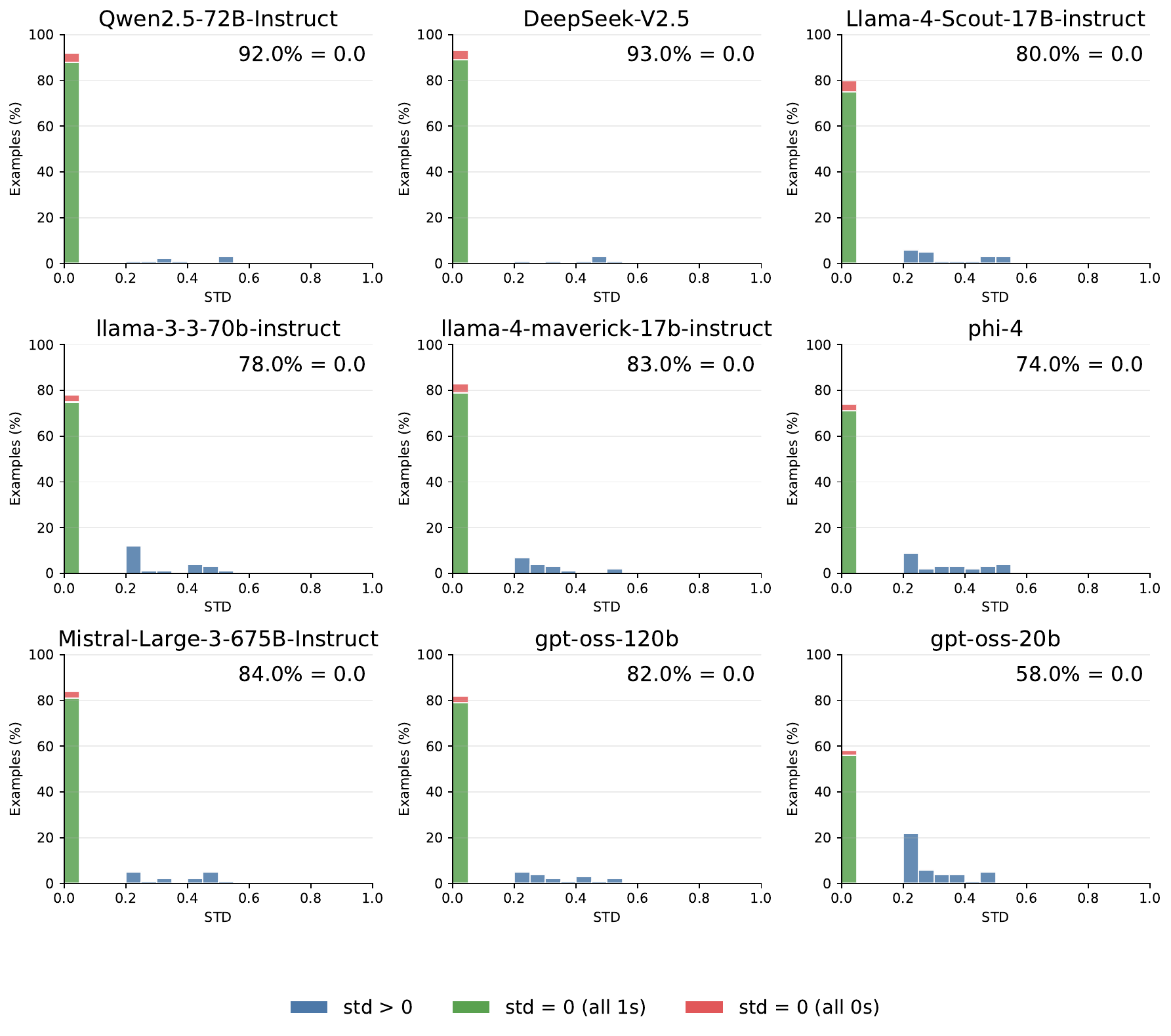}
    \caption{Per-example STD distribution for each model on the BoolQ dataset.}    
\end{figure*}

\begin{figure*}
    \centering
    \includegraphics[width=\textwidth]{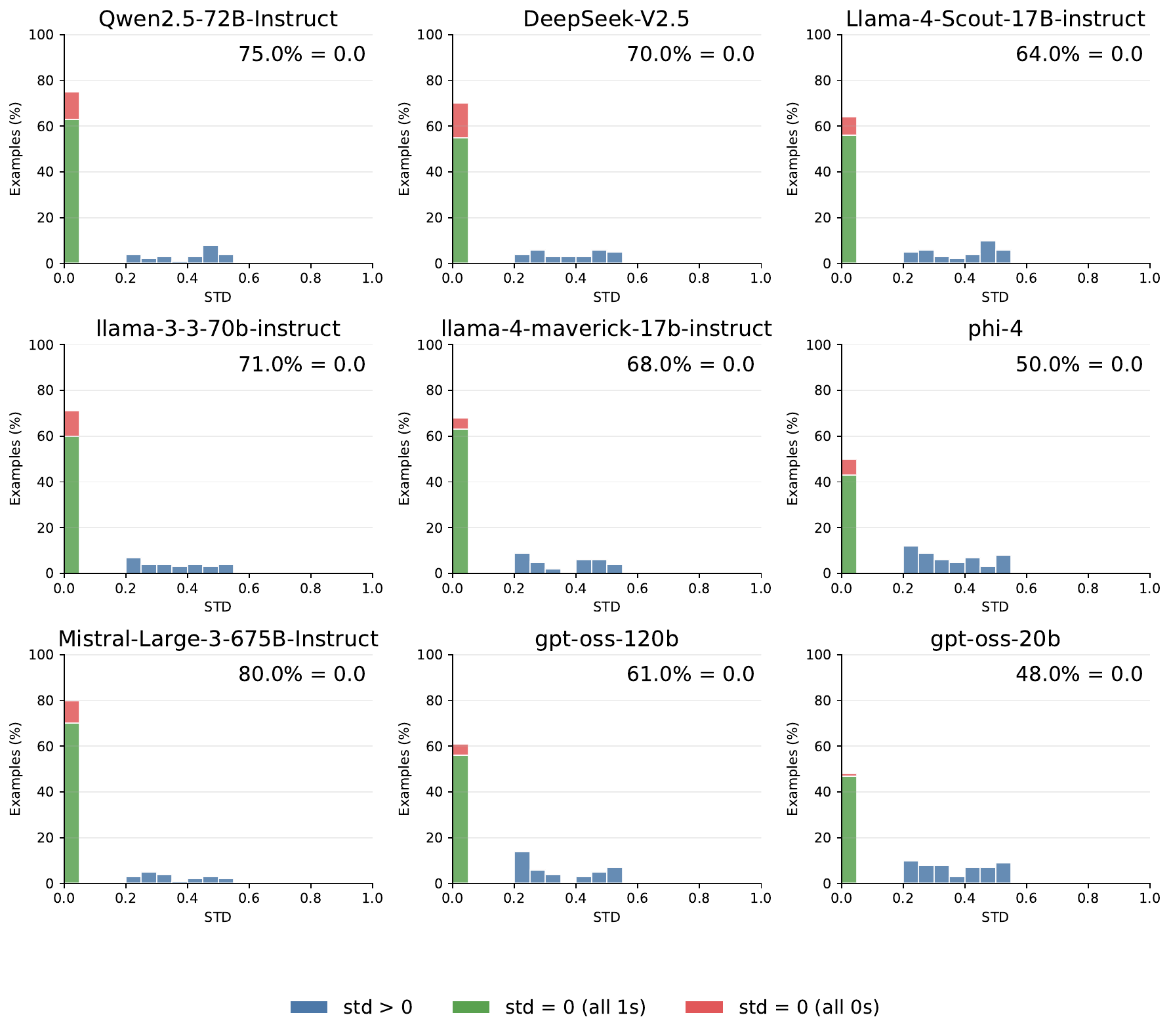}
    \caption{Per-example STD distribution for each model on the MMLU dataset.}    
\end{figure*}

\begin{figure*}
    \centering
    \includegraphics[width=\textwidth]{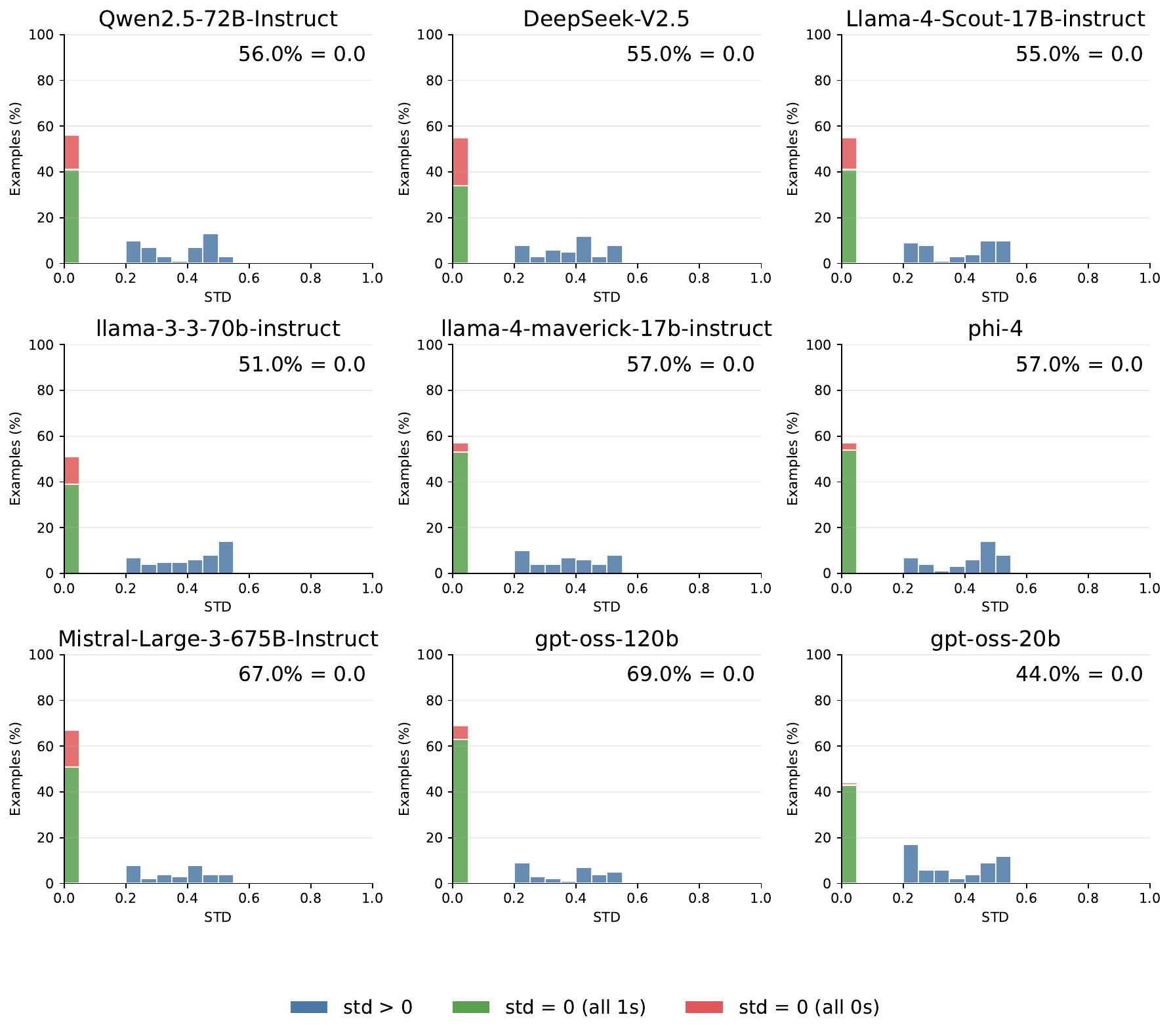}
    \caption{Per-example STD distribution for each model on the MMLU-Pro dataset.}    
\end{figure*}

\begin{figure*}
    \centering
    \includegraphics[width=\textwidth]{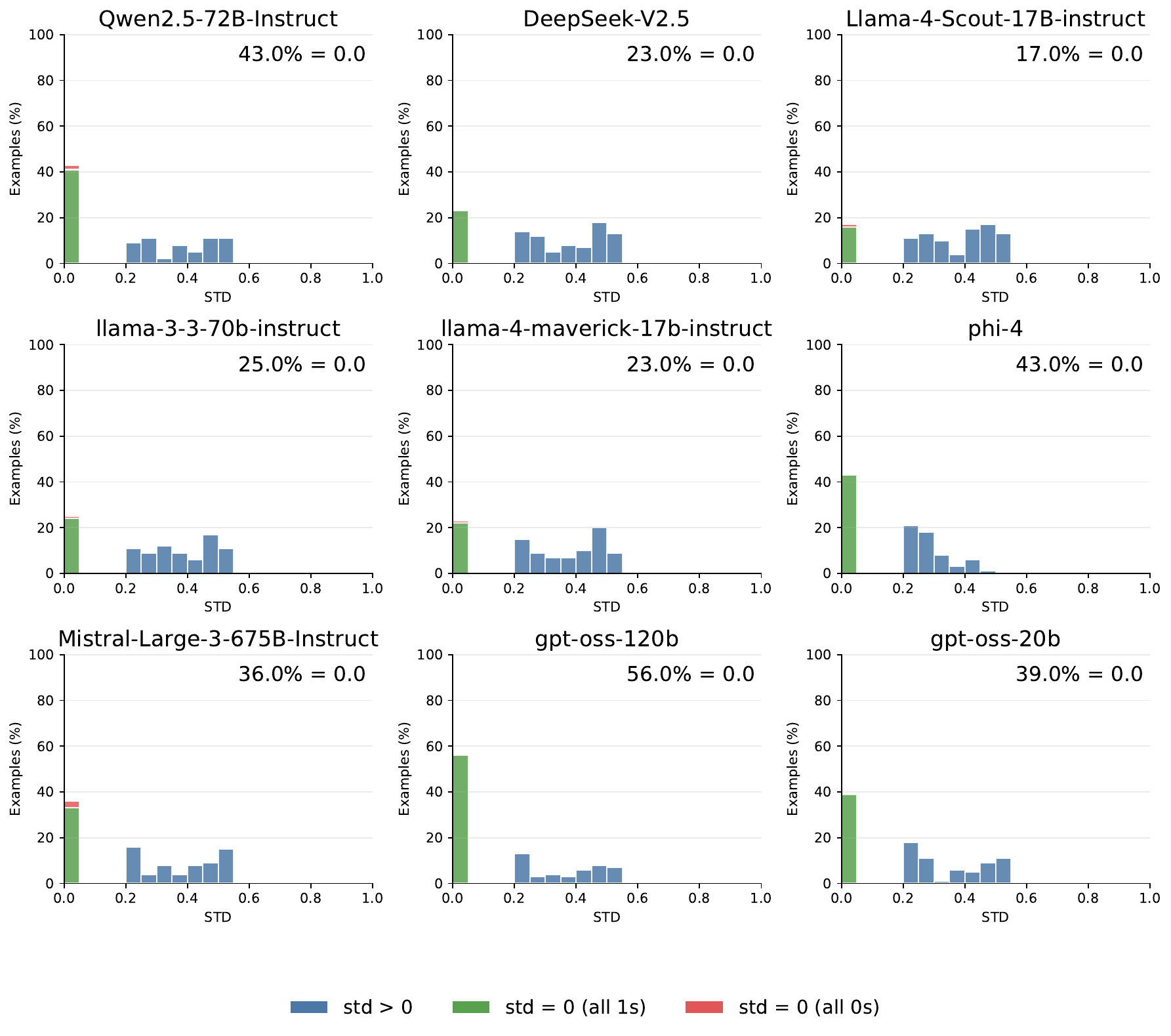}
    \caption{Per-example STD distribution for each model on the RewardBench dataset.}    
\end{figure*}

\begin{figure*}
    \centering
    \includegraphics[width=\textwidth]{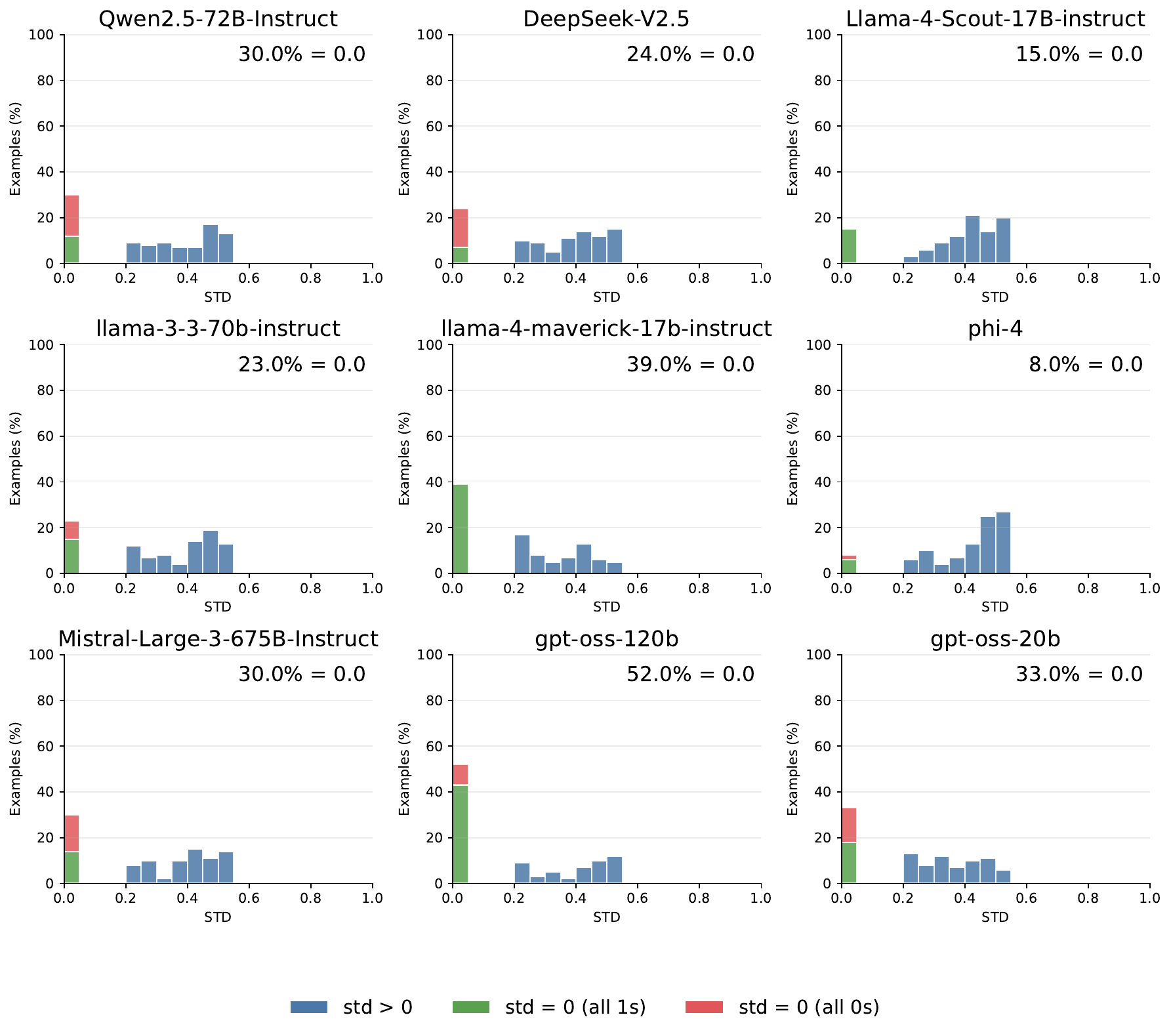}
    \caption{Per-example STD distribution for each model on the GPQA dataset.}    
\end{figure*}


\subsection{Statistical Testing}\label{app_sec:anova}

We analyzed the influence of four experimental factors (number of demos, prompt variation, template, and temperature) on performance across the datasets using both Type II and Type III ANOVA. Type II ANOVA evaluates each factor after accounting for all other factors, assuming no interaction terms, while Type III ANOVA additionally considers the presence of interactions and tests each factor after adjusting for all other factors and interactions. These tests assess whether different choices for each parameter significantly affect the results. 

Overall, the findings indicate that parameter choices have only a minor impact on performance. As shown in Table~\ref{tab:anova_summary_type2} and Table~\ref{tab:anova_summary_type3}, the majority of p-values (approximately $70\%$) are not significant. While prompt variation and number of demos exhibit statistically significant differences in some datasets (e.g., RewardBench, MMLU and MMLU-Pro), their effect sizes remain very small (<$0.005$), suggesting limited practical impact.

\begin{table*}
\begin{tabular}{lllrlllll}
\toprule
\toprule
dataset & factor & F & p-value & eta\_sq & partial\_eta\_sq & sum\_sq \\
\midrule
      BoolQ &        C(num\_demos) &  0.245 &     0.620 & 1.135e-05 &      1.137e-05 &     0.018 \\
      BoolQ & C(prompt\_variation) & 12.338 & 4.412e-06 &     0.001 &          0.001 &     1.848 \\
      BoolQ &    C(template\_used) &  2.289 &     0.130 & 1.059e-04 &      1.060e-04 &     0.171 \\
      BoolQ &       C(temprature) &  0.022 &     0.881 & 1.030e-06 &      1.032e-06 &     0.002 \\
       GPQA &        C(num\_demos) &  0.061 &     0.805 & 2.924e-06 &      2.924e-06 &     0.015 \\
       GPQA & C(prompt\_variation) &  1.279 &     0.278 & 1.224e-04 &      1.224e-04 &     0.633 \\
       GPQA &    C(template\_used) &  0.003 &     0.954 & 1.579e-07 &      1.579e-07 & 8.164e-04 \\
       GPQA &       C(temprature) &  0.583 &     0.445 & 2.785e-05 &      2.786e-05 &     0.144 \\
       IMDB &        C(num\_demos) &  0.123 &     0.726 & 6.015e-06 &      6.016e-06 &     0.004 \\
       IMDB & C(prompt\_variation) &  0.892 &     0.410 & 8.747e-05 &      8.748e-05 &     0.058 \\
       IMDB &    C(template\_used) &  0.128 &     0.721 & 6.275e-06 &      6.276e-06 &     0.004 \\
       IMDB &       C(temprature) &  0.947 &     0.331 & 4.641e-05 &      4.642e-05 &     0.031 \\
       MMLU &        C(num\_demos) &  4.836 &     0.028 & 2.236e-04 &      2.239e-04 &     0.869 \\
       MMLU & C(prompt\_variation) & 15.438 & 1.996e-07 &     0.001 &          0.001 &     5.547 \\
       MMLU &    C(template\_used) &  3.175 &     0.075 & 1.468e-04 &      1.470e-04 &     0.570 \\
       MMLU &       C(temprature) &  0.724 &     0.395 & 3.346e-05 &      3.352e-05 &     0.130 \\
   MMLU-Pro &        C(num\_demos) &  5.030 &     0.025 & 2.328e-04 &      2.329e-04 &     1.197 \\
   MMLU-Pro & C(prompt\_variation) &  6.235 &     0.002 & 5.771e-04 &      5.773e-04 &     2.968 \\
   MMLU-Pro &    C(template\_used) &  1.124 &     0.289 & 5.202e-05 &      5.206e-05 &     0.268 \\
   MMLU-Pro &       C(temprature) &  0.063 &     0.802 & 2.914e-06 &      2.916e-06 &     0.015 \\
RewardBench &        C(num\_demos) & 10.468 &     0.001 & 4.799e-04 &      4.824e-04 &     2.037 \\
RewardBench & C(prompt\_variation) & 35.940 & 3.686e-23 &     0.005 &          0.005 &    20.980 \\
RewardBench &    C(template\_used) &  2.003 &     0.157 & 9.184e-05 &      9.235e-05 &     0.390 \\
RewardBench &       C(temprature) &  1.439 &     0.230 & 6.595e-05 &      6.632e-05 &     0.280 \\
\bottomrule
\bottomrule
\end{tabular}
\caption{Summary across datasets — Type II ANOVA.}
\label{tab:anova_summary_type2}
\end{table*}

\begin{table*}
\footnotesize
\begin{tabular*}{\textwidth}{@{\extracolsep{\fill}} ll lr lllll}
\toprule
\toprule
dataset & factor & F & p-value & eta\_sq & partial\_eta\_sq & sum\_sq \\
\midrule
      BoolQ &        C(num\_demos, Sum) &      0.245 &     0.620 & 9.265e-07 &      1.137e-05 &     0.018 \\
      BoolQ & C(prompt\_variation, Sum) &     12.338 & 4.412e-06 & 9.319e-05 &          0.001 &     1.848 \\
      BoolQ &    C(template\_used, Sum) &      2.289 &     0.130 & 8.645e-06 &      1.060e-04 &     0.171 \\
      BoolQ &       C(temprature, Sum) &      0.022 &     0.881 & 8.410e-08 &      1.032e-06 &     0.002 \\
      BoolQ &                Intercept & 243192.586 &     0.000 &     0.918 &          0.918 & 18209.710 \\
       GPQA &        C(num\_demos, Sum) &      0.061 &     0.805 & 1.615e-06 &      2.924e-06 &     0.015 \\
       GPQA & C(prompt\_variation, Sum) &      1.279 &     0.278 & 6.761e-05 &      1.224e-04 &     0.633 \\
       GPQA &    C(template\_used, Sum) &      0.003 &     0.954 & 8.723e-08 &      1.579e-07 & 8.164e-04 \\
       GPQA &       C(temprature, Sum) &      0.583 &     0.445 & 1.539e-05 &      2.786e-05 &     0.144 \\
       GPQA &                Intercept &  16936.111 &     0.000 &     0.447 &          0.447 &  4187.914 \\
       IMDB &        C(num\_demos, Sum) &      0.123 &     0.726 & 2.021e-07 &      6.016e-06 &     0.004 \\
       IMDB & C(prompt\_variation, Sum) &      0.892 &     0.410 & 2.939e-06 &      8.748e-05 &     0.058 \\
       IMDB &    C(template\_used, Sum) &      0.128 &     0.721 & 2.108e-07 &      6.276e-06 &     0.004 \\
       IMDB &       C(temprature, Sum) &      0.947 &     0.331 & 1.559e-06 &      4.642e-05 &     0.031 \\
       IMDB &                Intercept & 586755.026 &     0.000 &     0.966 &          0.966 & 18989.883 \\
       MMLU &        C(num\_demos, Sum) &      4.836 &     0.028 & 5.261e-05 &      2.239e-04 &     0.869 \\
       MMLU & C(prompt\_variation, Sum) &     15.438 & 1.996e-07 & 3.359e-04 &          0.001 &     5.547 \\
       MMLU &    C(template\_used, Sum) &      3.175 &     0.075 & 3.454e-05 &      1.470e-04 &     0.570 \\
       MMLU &       C(temprature, Sum) &      0.724 &     0.395 & 7.873e-06 &      3.352e-05 &     0.130 \\
       MMLU &                Intercept &  70297.530 &     0.000 &     0.765 &          0.765 & 12630.152 \\
   MMLU-Pro &        C(num\_demos, Sum) &      5.030 &     0.025 & 9.103e-05 &      2.329e-04 &     1.197 \\
   MMLU-Pro & C(prompt\_variation, Sum) &      6.235 &     0.002 & 2.257e-04 &      5.773e-04 &     2.968 \\
   MMLU-Pro &    C(template\_used, Sum) &      1.124 &     0.289 & 2.034e-05 &      5.206e-05 &     0.268 \\
   MMLU-Pro &       C(temprature, Sum) &      0.063 &     0.802 & 1.140e-06 &      2.916e-06 &     0.015 \\
   MMLU-Pro &                Intercept &  33643.358 &     0.000 &     0.609 &          0.609 &  8007.085 \\
RewardBench &        C(num\_demos, Sum) &     10.468 &     0.001 & 4.018e-04 &      4.824e-04 &     2.037 \\
RewardBench & C(prompt\_variation, Sum) &     35.940 & 3.686e-23 &     0.004 &          0.005 &    20.980 \\
RewardBench &    C(template\_used, Sum) &      2.003 &     0.157 & 7.689e-05 &      9.235e-05 &     0.390 \\
RewardBench &       C(temprature, Sum) &      1.439 &     0.230 & 5.522e-05 &      6.632e-05 &     0.280 \\
RewardBench &                Intercept &   4240.655 &     0.000 &     0.163 &          0.164 &   825.159 \\
\bottomrule
\bottomrule
\end{tabular*}
\caption{Summary across datasets — Type III ANOVA.}
\label{tab:anova_summary_type3}
\end{table*}

\newpage\null
\newpage\null
\newpage\null

\begin{figure*}[b]
    \centering
     \includegraphics[width=\textwidth]{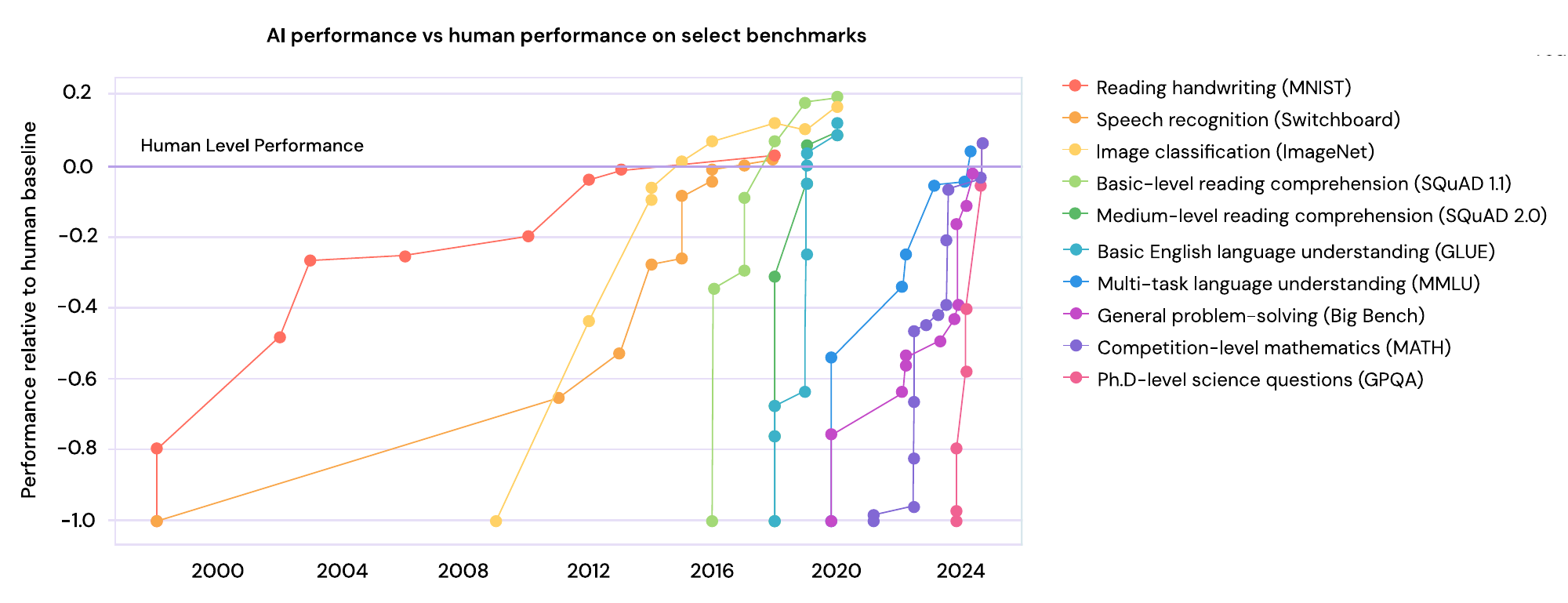}
    \caption*{A figure published as part of the International AI Safety Report \citep{bengio2025internationalaisafetyreport} demonstrating the rapid advance of AI model performance from 1998 to 2024. In recent benchmarks, models progressed quickly from poor performance to surpassing human experts.}
    \label{fig:saturate_prog}
\end{figure*}

\newpage

\section{Saturation Progress}\label{app:saturation_prog}
Saturation in LLM evaluation is a well-known challenge, with numerous papers highlighting, analyzing, or accounting for it in their studies. Here, we provide evidence from existing research showing that saturation is becoming increasingly prevalent. This strengthens our findings: as models improve in capabilities and benchmarks lose relevance, their ability to generalize also becomes more critical.

\begin{figure*}
    \centering
    \includegraphics[width=.9\textwidth]{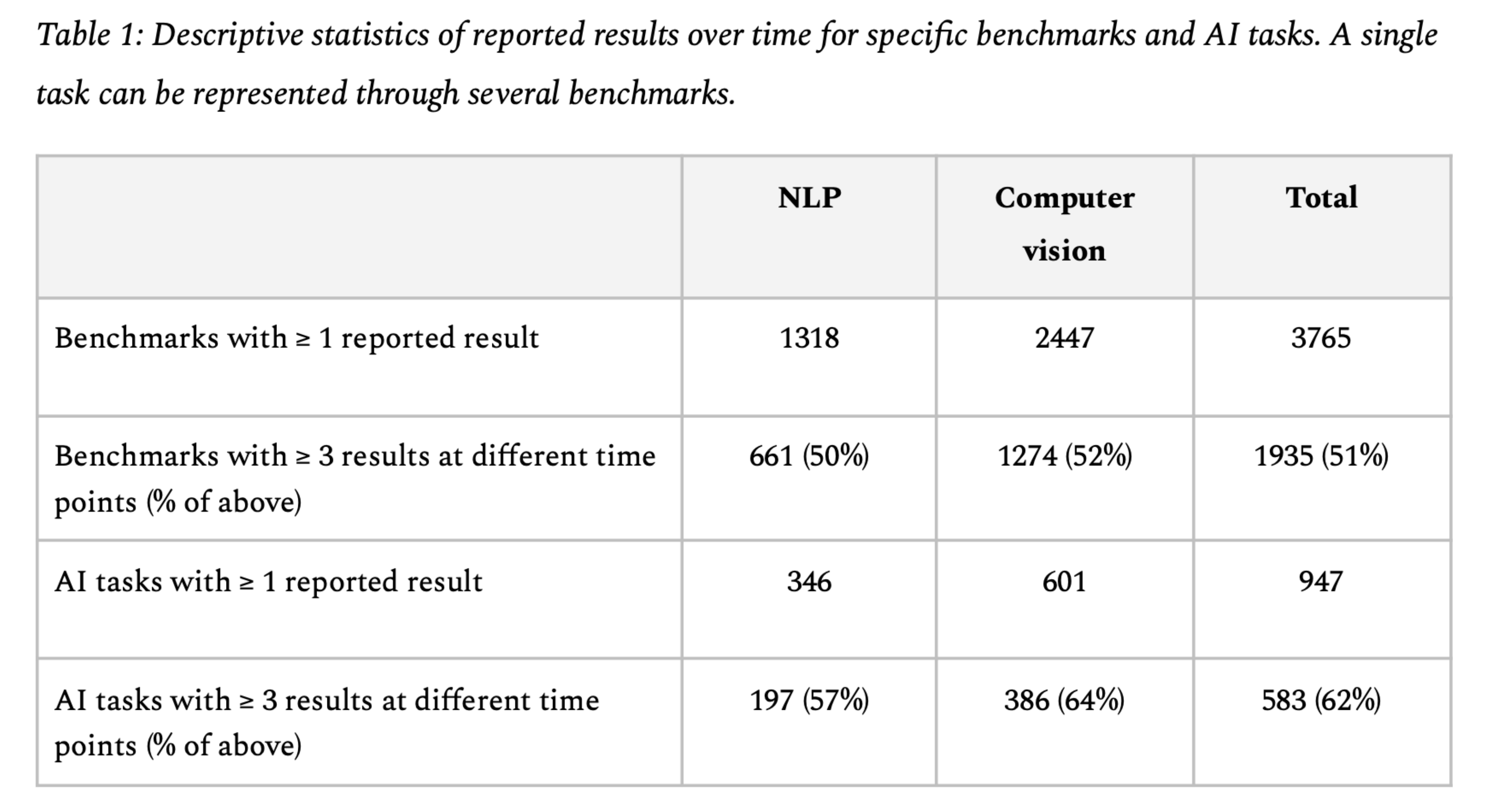}
    \caption{A table taken from \cite{Ott2022MappingGD} that demonstrates how, over time, the number of benchmarks relevant for repeated reporting drops significantly. For example, only $50\%$ of NLP benchmarks have results at three or more time points.}
    \label{fig:placeholder6}
\end{figure*}
\begin{figure*}
    \centering
    \includegraphics[width=.8\textwidth]{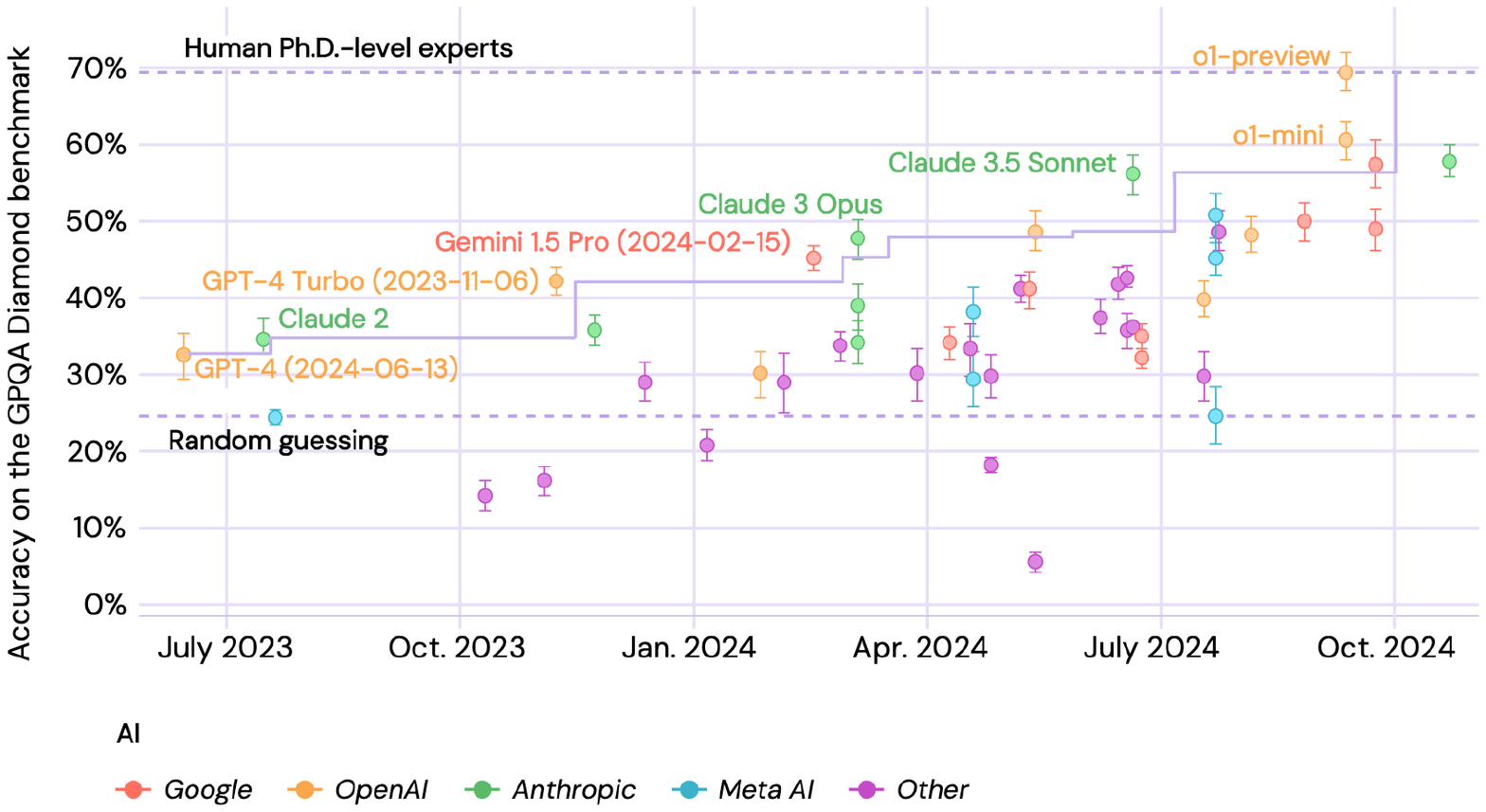}
    \caption{A figure published as part of the International AI Safety Report \citep{bengio2025internationalaisafetyreport} demonstrating that AI models improved from near-random (33\%) to expert-level (70\%) accuracy on PhD-level science questions within 15 months,}
    \label{fig:placeholder7}
\end{figure*}





\null\newpage
\null\newpage
\section{AI Assistance Usage}
We used AI for paraphrasing and improving clarity of writing only. 

\end{document}